\documentclass{article}

\PassOptionsToPackage{numbers, compress}{natbib}

\usepackage{colortbl}

 \usepackage[preprint]{neurips_2025}
\usepackage{amsmath}
\usepackage{graphicx}
\usepackage{subcaption}
\usepackage{wrapfig}
\usepackage{tabularx}

\usepackage[utf8]{inputenc} 
\usepackage[T1]{fontenc}    
\usepackage{hyperref}       
\usepackage{url}            
\usepackage{booktabs}       
\usepackage{amsfonts}       
\usepackage{nicefrac}       
\usepackage{microtype}      
\usepackage{xcolor}         
\usepackage{graphicx}
\usepackage{pifont}  
\usepackage{tablefootnote}
\newcommand{\cmark}{\ding{51}} 
\newcommand{\xmark}{\phantom{\ding{51}}}

\title{Scalable Chain of Thoughts via Elastic Reasoning}

\author{
Yuhui Xu\quad Hanze Dong\quad Lei Wang\quad Doyen Sahoo\quad Junnan Li\quad Caiming Xiong\\\\ Salesforce AI Research
}

\begin{document}

\maketitle

\begin{abstract}
Large reasoning models (LRMs) have achieved remarkable progress on complex tasks by generating extended chains of thought (CoT). However, their uncontrolled output lengths pose significant challenges for real-world deployment, where inference-time budgets on tokens, latency, or compute are strictly constrained. We propose \textbf{Elastic Reasoning}, a novel framework for scalable chain of thoughts that explicitly separates reasoning into two phases—\textit{thinking} and \textit{solution}—with independently allocated budgets. At test time, Elastic Reasoning prioritizes the completeness of solution segments, significantly improving reliability under tight resource constraints. To train models that are robust to truncated thinking, we introduce a lightweight \textit{budget-constrained rollout} strategy, integrated into GRPO, which teaches the model to reason adaptively when the thinking process is cut short and generalizes effectively to unseen budget constraints without additional training. Empirical results on mathematical (AIME, MATH500) and programming (LiveCodeBench, Codeforces) benchmarks demonstrate that Elastic Reasoning performs robustly under strict budget constraints, while incurring significantly lower training cost than baseline methods. Remarkably, our approach also produces more concise and efficient reasoning even in unconstrained settings. Our code has been made
available at \url{https://github.com/SalesforceAIResearch/Elastic-Reasoning}.

\end{abstract}

\section{Introduction}\label{sec:intro}

Large reasoning models (LRMs)~\citep{deepseekai2025deepseekr1incentivizingreasoningcapability,openai2024openaio1card} have demonstrated remarkable performance on complex reasoning tasks by producing extended Chain-of-Thought (CoT) outputs, which facilitate effective problem-solving in domains such as mathematics and programming. Reinforcement learning (RL) techniques \citep{schulman2017proximal,zelikman2022star,rafailov2023direct,dong2023raft,shao2024deepseekmath}, have been employed to optimize these reasoning trajectories, enabling LRMs to generate longer, more informative chains. These RL-driven methods scale effectively across diverse benchmarks \citep{zhang2024o1,dong2024rlhf,deepscaler2025,xiong2025self,deepcoder2025}, yielding substantial gains in both solution accuracy and robustness; while they often incur significantly longer inference chains~\citep{deepseekai2025deepseekr1incentivizingreasoningcapability,du2025virgo,yu2024distilling,qin2024o1,xiong2025minimalist}. Notably, the length of the reasoning trajectory remains uncontrolled, making it difficult to allocate a fixed compute budget at inference time while maintaining a desired performance level.

Two primary lines of research have been proposed to address this challenge. The first, known as \textbf{Long2Short}~\cite{team2501kimi,kang2024c3otgeneratingshorterchainofthought}, seeks to reduce reasoning length through reinforcement learning with trajectory penalties or compression-aware fine-tuning, where the model is trained on shortened trajectories to preserve performance while minimizing inference cost. The second line of work focuses on \textbf{length control}~\citep{muennighoff2025s1simpletesttimescaling,aggarwal2025l1,yuan2024followinglengthconstraintsinstructions}. S1~\citep{muennighoff2025s1simpletesttimescaling} introduces a simple mechanism that prompts the model to emit special tokens (e.g., ``Wait'', ``Final Answer'') to regulate reasoning length. However, this approach significantly degrades performance, as it overlooks the critical role of the solution segment. L1~\cite{aggarwal2025l1} proposes a reinforcement learning framework that enforces explicit length constraints over the entire trajectory. While more flexible, this method demands substantial training resources and still results in noticeable performance degradation compared to the original model.

\begin{wrapfigure}{r}{0.5\textwidth}
\vspace{-2em}
    \centering
    \includegraphics[width=0.5\textwidth]{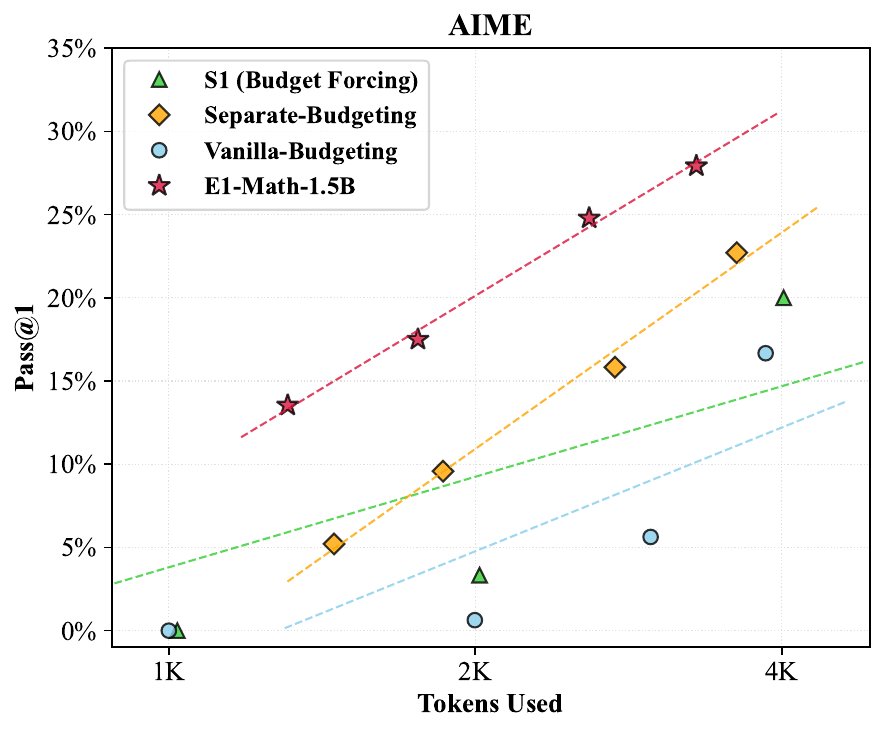}
    \caption{Separating thinking and solution phases enables better length control.}
    \label{fig:motivation}
    \vspace{-2em}
\end{wrapfigure}

We propose \textbf{Elastic Reasoning}, a simple yet effective method that enables large reasoning models to achieve scalable and adaptive length control. As illustrated in Figure~\ref{fig:motivation}, the S1 approach---generating the answer by emitting a special token such as ``Final Answer''---performs better than directly truncating the full reasoning trajectory, underscoring the importance of preserving the solution segment. Motivated by this, we propose separate budgeting which explicitly divides the total token budget \( c \) into two parts: \( t \) tokens for the \emph{thinking} phase and \( s \) tokens for the \emph{solution} phase, where \( c = t + s \). Once the model consumes \( t \) tokens in the thinking phase, we forcibly terminate it by appending the special token \texttt{</think>} and transition to solution generation. Separate budgeting outperforms S1 under varying generation budgets.

To further improve solution quality under incomplete reasoning, we introduce a novel training strategy called \emph{budget-constrained rollout}, which teaches the model to generate high-quality answers even with partial CoT trajectories. This method is integrated into GRPO training and is highly efficient—requiring only 200 training steps on math tasks with a maximum response length of 2K tokens (\( t^* = 1\text{K},\ s^* = 1\text{K} \)), compared to 700 steps for L1-Exact and 820 steps for L1-Max with a 4K response length. Moreover, models trained with Elastic Reasoning generalize effectively to arbitrary reasoning budgets without the need for further fine-tuning.

We evaluate Elastic Reasoning on both mathematical and programming reasoning tasks, introducing two models: \textbf{E1-Math-1.5B} and \textbf{E1-Code-14B}. (1) \textbf{E1-Math-1.5B} outperforms both L1-Exact and S1, and achieves performance comparable to L1-Max, while requiring significantly fewer training steps. For instance, on the AIME2024 dataset, our method achieves 35.0\% accuracy, compared to 27.1\% for L1-Max, 24.2\% for L1-Exact, and 41.0\% for the original model. (2) \textbf{E1-Code-14B} demonstrates strong scaling with varying inference budgets, achieving a Codeforces rating of 1987 and placing in the 96.0 percentile—comparable to O1-2024-12-17 (Low), which scores 1991 and ranks in the 96.1 percentile. (3) A surprising observation is that, after training, the trajectories generated by our models are significantly shorter than those from the original DeepScaleR and DeepCoder models across both math and code tasks. This suggests that budget-constrained rollout not only improves length control but also encourages the model to reason more concisely and generate more efficient solutions.

\section{Related works}\label{sec:related}

\subsection{Test-time scaling in large language models}
Increasing computation during inference, often referred to as test-time scaling (TTS), has been shown to improve the reasoning capabilities of LLMs~\citep{wei2023chainofthoughtpromptingelicitsreasoning, wang2023selfconsistencyimproveschainthought, snell2024scalingllmtesttimecompute, deepseekai2025deepseekr1incentivizingreasoningcapability, team2501kimi, muennighoff2025s1simpletesttimescaling}. 
Early works, such as chain-of-thought prompting~\citep{wei2023chainofthoughtpromptingelicitsreasoning}, show that producing a series of intermediate reasoning steps significantly improves LLMs’ performance on complex reasoning tasks. 
Building on this, self-consistency~\citep{wang2023selfconsistencyimproveschainthought} further boosts performance by sampling a diverse set of reasoning paths and selecting the most consistent answer.
Recent studies have formalized these findings into test-time inference scaling laws~\citep{snell2024scalingllmtesttimecompute, wu2024inferencescalinglawsempirical}.
~\citet{wu2024inferencescalinglawsempirical} explore the trade-offs between model size and inference-time computation.
~\citet{snell2024scalingllmtesttimecompute} investigated how fixed but non-trivial inference-time budgets can significantly boost LLM performance.
The remarkable successes of advanced reasoning models, such as o1~\citep{openai2024openaio1card} and R1~\citep{deepseekai2025deepseekr1incentivizingreasoningcapability}, have further amplified interest in leveraging TTS techniques.
While much of the existing works primarily focuses on improving performance by increasing inference-time computation, our work takes a different perspective: \textit{How can we enable LLMs to perform effective long reasoning under strict output length constraints?}

\subsection{Length control in large language models}

Controlling the generation length of an LLM directly affects both latency and monetary cost at inference time.
Earlier approaches to length control are designed mainly for general text generation~\citep{jie2023prompt, yuan2024followinglengthconstraintsinstructions}. Typical methods include (i) manipulating positional encodings to achieve exact sequence lengths~\citep{butcher2024preciselengthcontrollarge}, (ii) modifying training objectives to penalize deviations from length targets~\citep{jie2023prompt, singhal2024longwaygoinvestigating}, and (iii) fine-tuning on instructions that explicitly state the desired output length~\citep{yuan2024followinglengthconstraintsinstructions}.
Although effective for tasks such as summarization or constrained writing, these techniques generally aim to verbosity or enforce maximum-length limits, and overlook the intricate, step-by-step reasoning processes required for many reasoning tasks.
Recent works have begun to explore efficiency in reasoning by encouraging shorter chains~\citep{kang2024c3otgeneratingshorterchainofthought, arora2025traininglanguagemodelsreason}; however, they typically lack mechanisms for precise, user-defined length targets that align with explicit compute budgets. One notable attempt, budget forcing~\citep{muennighoff2025s1simpletesttimescaling}, enforces strict token caps by truncating or padding with special tokens. This can yield incomplete reasoning or unnatural, forced outputs, ultimately harming both accuracy and interpretability.
Additionally, L1~\citep{aggarwal2025l1} uses reinforcement learning to let models dynamically allocate inference compute based on constraints provided in the prompt.
Our approach does not need to include length instructions in the prompt. Instead, we truncate reasoning trajectories to meet a given budget and train the model under these constraints via reinforcement learning.

\subsection{Efficient reasoning in large language models}
Making complex reasoning in LLMs more efficient, particularly by shortening the reasoning process, is crucial to reducing computational costs and making these models practical for real-world deployment.
This has become a vibrant research area with several promising directions to encourage more concise and effective reasoning strategies~\citep{kang2024c3otgeneratingshorterchainofthought,xu2024think, hao2024training, liao2025reward, luo2025o1}.
One common strategy involves incorporating explicit rewards into RL to encourage the model to find shorter reasoning paths~\citep{team2501kimi, luo2025o1}.
Some focus on creating datasets with examples of concise reasoning paths and then using SFT teach models how to generate compact and knowledgeable reasoning steps~\citep{kang2024c3otgeneratingshorterchainofthought, yu2024distilling}.
Instead of relying solely on explicit textual reasoning, methods exploring latent reasoning aim to compress these intermediate steps into more compact, internal representations~\citep{hao2024training, shen2025efficient, saunshi2025reasoning}.
Efficiency can also be improved during inference, without needing to retrain the model. These training-free techniques dynamically adapt the reasoning strategy based on the specific input or task demands~\citep{liao2025reward, fu2025reasoning}.
In this work, we introduce a training approach using reinforcement learning under strict budget constraints to encourage the model to balance reasoning quality with cost efficiency.

\section{Methodology}\label{sec:method}

\begin{figure}
    \centering
    \includegraphics[width=0.95\linewidth]{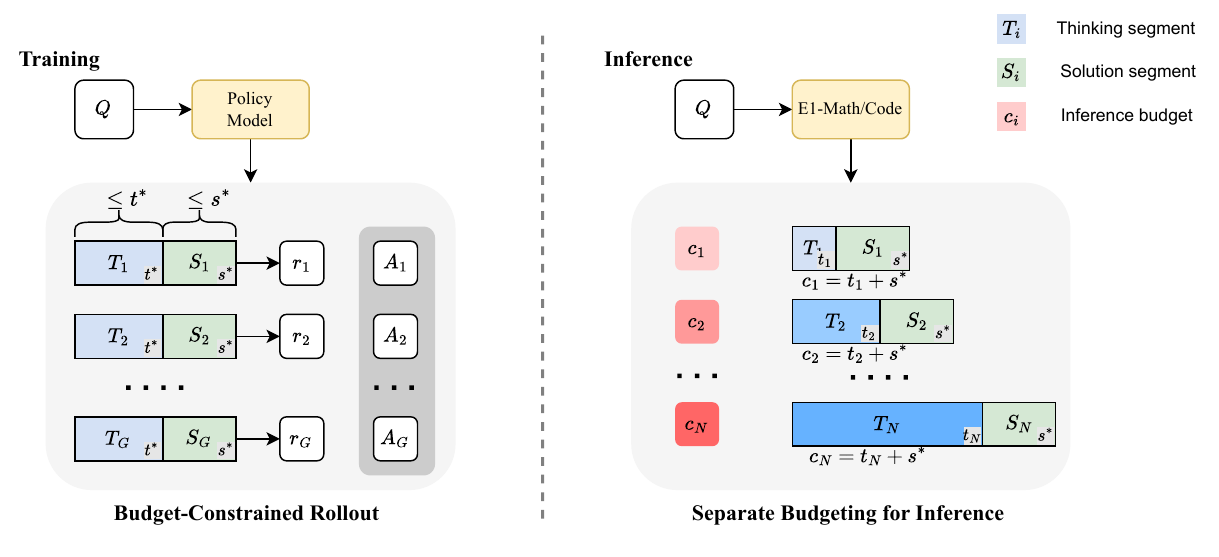}
    \caption{The framework of Elastic Reasoning. Elastic Reasoning comprises two key components: (1) GRPO training with budget-constrained rollout, and (2) separate budgeting for inference. \textbf{Left:} During training, the model is optimized using GRPO under a fixed token budget \((t^*, s^*)\). \textbf{Right:} At inference time, the trained E1 model can generalize to arbitrary token budgets \(c_i = t_i + s^*\), enabling flexible and efficient reasoning.}
    \label{fig:framework}
\end{figure}

\subsection{Preliminaries: Reasoning Language Models}

We consider reasoning-augmented language models that generate outputs consisting of two distinct segments: a \emph{thinking} part and a \emph{solution} part. Following prior work, we denote the reasoning phase using special tokens such as \texttt{<think>} and \texttt{</think>} to explicitly mark the model’s intermediate thoughts.

Formally, given an input prompt \( x \), the model generates an output sequence \( y = (y^{\text{think}}, y^{\text{solution}}) \), where \( y^{\text{think}} \) contains the intermediate reasoning steps (enclosed between \texttt{<think>} and \texttt{</think>}) and \( y^{\text{solution}} \) contains the final solution. Typically, \( y^{\text{think}} \) accounts for over 90\% of the total tokens, while \( y^{\text{solution}} \) provides a concise summary and final answer. The overall generation structure is:
\[
y = (\texttt{<think>} \; \text{intermediate reasoning} \; \texttt{</think>}, \; \text{solution})
\]

\subsection{Elastic reasoning}

\subsubsection{Budget-constrained inference}

In many real-world applications, inference cost must be carefully controlled due to constraints on latency, computation, or memory. A common approach is to truncate generation after a fixed number of tokens \( c \), enforcing:
\[
|y| \leq c
\]
where \( |y| \) denotes the number of generated tokens. However, naively truncating the output often results in incomplete or missing \( y^{\text{solution}} \), leading to invalid or unusable predictions.

\subsubsection{Separate budgeting for thinking and solution}

To address this limitation, we propose \emph{Separate Budgeting}, a method that explicitly allocates independent budgets for the reasoning and solution phases. A key observation is that even when the reasoning phase is forcibly terminated (e.g., by inserting \texttt{</think>}), the model is still capable of producing a coherent—and often correct—solution.

Given a total generation budget \( c \), we divide it into two components: a budget \( t \) for the thinking phase and a budget \( s \) for the solution phase, such that \( c = t + s \).

During inference:
\begin{itemize}
    \item The model begins generating within a \texttt{<think>} block.
    \item If the model emits \texttt{</think>} before reaching the budget \( t \), we transition immediately to the solution phase.
    \item If the budget \( t \) is exhausted before \texttt{</think>} is emitted, we forcibly terminate the reasoning by appending \texttt{</think>}.
    \item The model then continues generating the solution segment, up to a maximum of \( s \) tokens.
\end{itemize}

This approach ensures that both the reasoning and solution components are explicitly accommodated within the total budget \( c \), thereby avoiding unintended truncation of the solution segment. The thinking budget \( t \) can be flexibly adjusted at inference time to match different application scenarios, while the solution phase always retains a guaranteed allocation. As shown in Figure~\ref{fig:motivation}, Separate Budgeting outperforms both vanilla budgeting (naïve truncation) and S1 (budget forcing). By dedicating a fixed token budget for solution generation, Separate Budgeting significantly improves the reliability and quality of model outputs under tight inference-time constraints.

\subsubsection{Budget-constrained rollout}

While Separate Budgeting ensures dedicated budgets for both reasoning and solution phases, we observe that naively truncating the thinking part—especially on complex tasks such as code generation—can lead to significant performance degradation. To mitigate this issue, we propose a reinforcement learning (RL) fine-tuning procedure that explicitly trains the model under reasoning budget constraints, allowing it to produce more effective and concise reasoning within limited budgets.

We adopt GRPO as our RL algorithm. Let \( \pi_\theta \) denote the policy of a language model parameterized by \( \theta \), which generates a response \( y = (y^{\text{think}}, y^{\text{solution}}) \) for a given input \( x \), subject to a total budget constraint \( t^* + s^* = c^* \). During training, we simulate the Separate Budgeting procedure used at inference time: the policy rolls out a reasoning segment \( y^{\text{think}} \) up to a maximum of \( t^* \) tokens. If the model emits the \texttt{</think>} token before reaching this limit, it proceeds to generate the solution segment as usual. Otherwise, we forcibly append \texttt{</think>} once the budget \( t^* \) is reached. The model then generates the solution segment \( y^{\text{solution}} \) using the remaining \( s^* \) tokens.

Let \( r(y) \) denote a task-specific reward function. The training objective is to maximize the expected reward:
\[
J(\theta) = \mathbb{E}_{x \sim \mathcal{D},\ y \sim \pi_\theta(\cdot \mid x;\ t^*, s^*)} \left[ r(y) \right]
\]

We optimize \( J(\theta) \) using GRPO with the following gradient estimator:
\[
\nabla_\theta J(\theta) = \mathbb{E}_{x, y} \left[ A(x, y) \nabla_\theta \log \pi_\theta(y \mid x;\ t^*, s^*) \right]
,\quad
A(x, y) = \frac{r(y) - \mathbb{E}_{y' \sim \pi_\theta(\cdot \mid x;\ t^*, s^*)}[r(y')]}{\sqrt{\mathbb{V}_{y' \sim \pi_\theta(\cdot \mid x)}[r(y')]}}
\]

In our training setup, we fix the budget pair to \( (t^*, s^*) = (1\text{K}, 1\text{K}) \) for simplicity and efficiency. Surprisingly, we find that the learned policy generalizes well to a wide range of unseen budget configurations at test time, without requiring any additional fine-tuning. As shown in Figure~\ref{fig:motivation}, the E1-Math-1.5B model achieves substantial improvements while generalizing robustly across various generation budgets. This indicates that Elastic Reasoning encourages the model to internalize a flexible reasoning strategy that adapts to different resource constraints.

This RL-based adaptation helps the model prioritize informative reasoning content earlier in the generation process, thereby improving both robustness and solution quality under test-time truncation.

\begin{figure}[t]
  \centering
  \begin{subfigure}[b]{0.48\textwidth}
      \includegraphics[width=\textwidth]{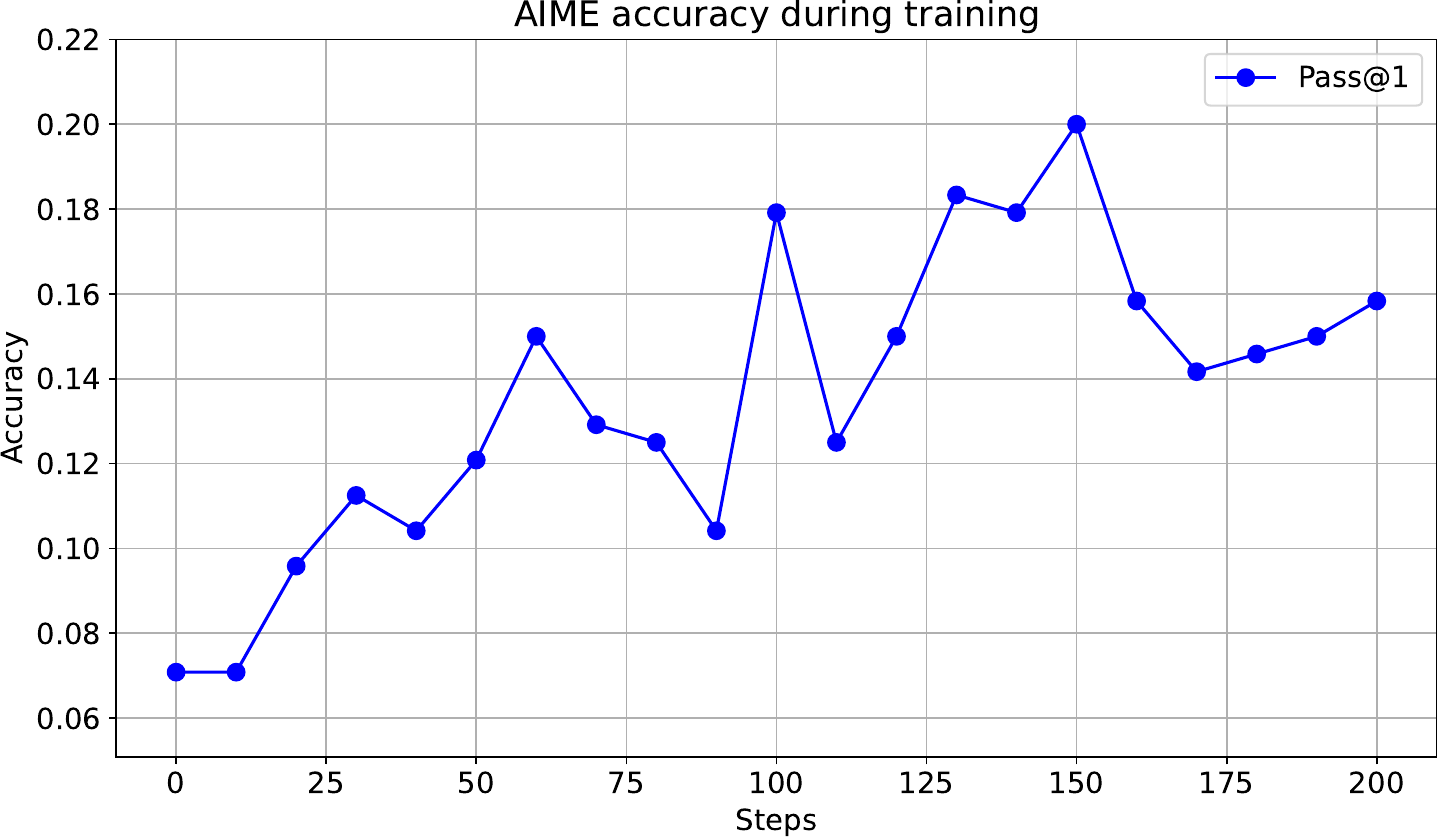}
      \caption{Pass@1 on AIME2024 over steps}
  \end{subfigure}
  \hfill
  \begin{subfigure}[b]{0.48\textwidth}
      \includegraphics[width=\textwidth]{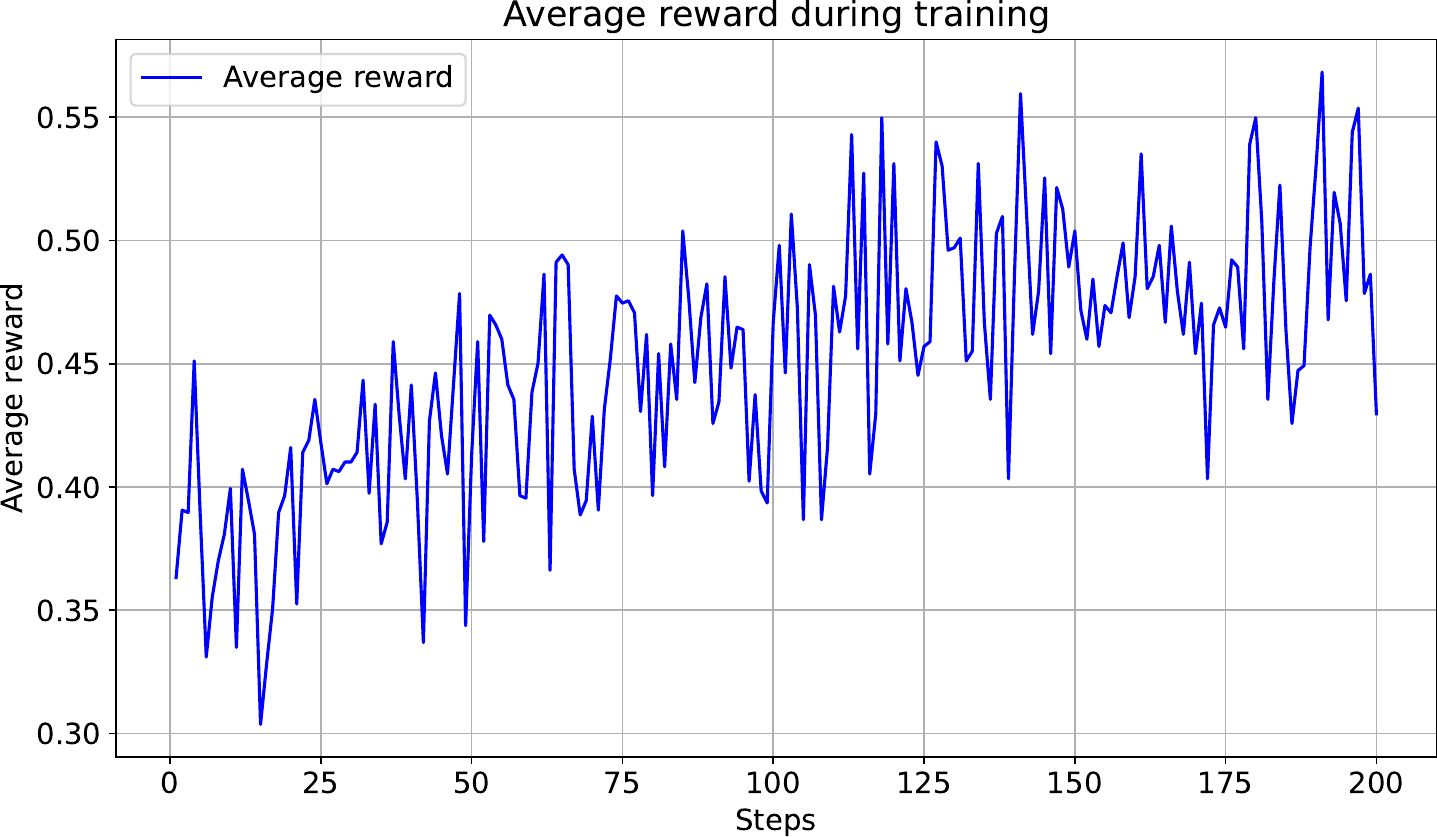}
      \caption{Reward over steps}
  \end{subfigure}
  \caption{Validation accuracy and reward curves of E1-Math-1.5B over training steps.}
  \label{fig:reward}
\end{figure}

\section{Experiment results}\label{sec:experiment}
\begin{figure}
    \centering
    \includegraphics[width=0.95\linewidth]{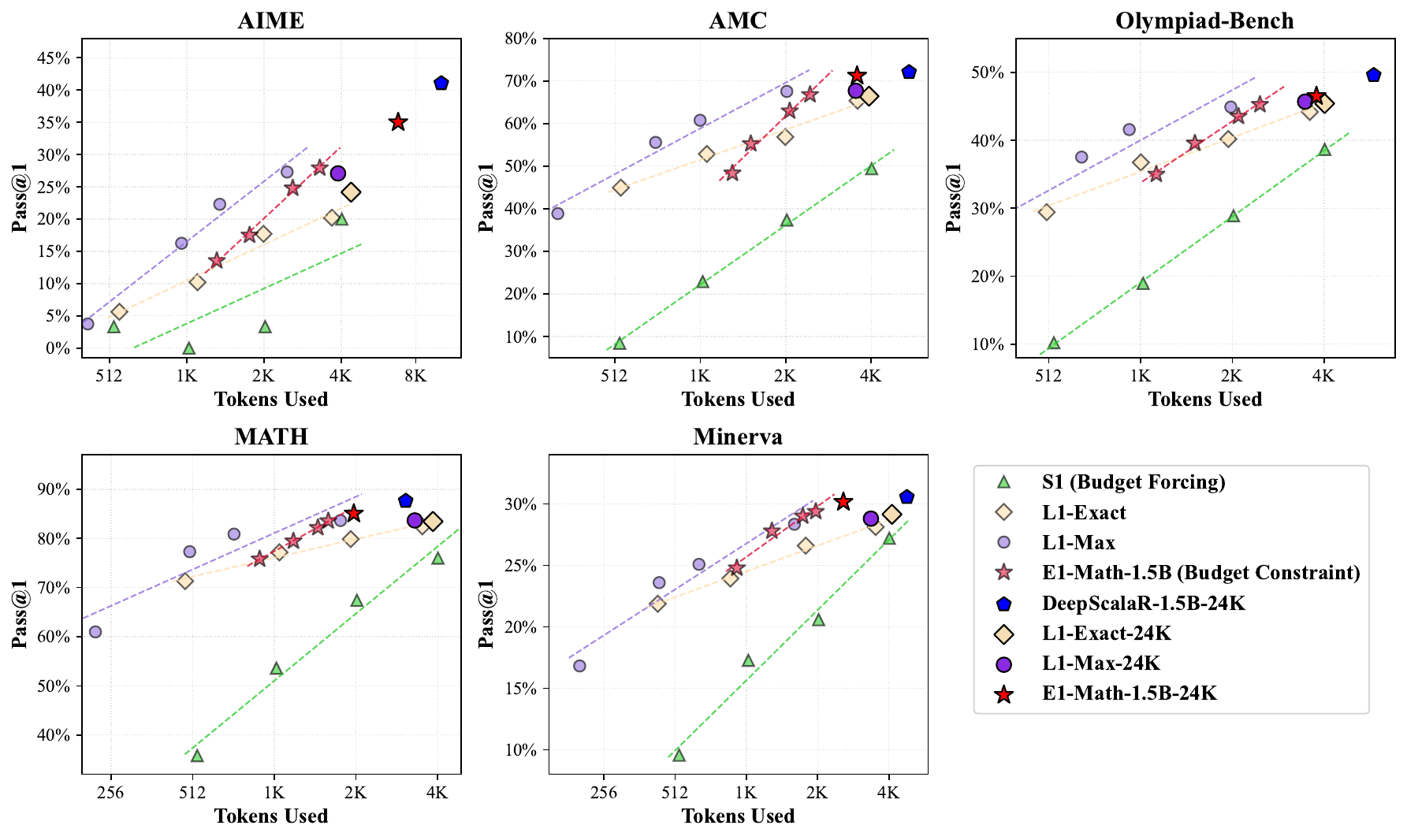}
    \caption{Comparison of E1-Math-1.5B with L1 and S1 baselines under varying generation budgets.}
    \label{fig:math}
\end{figure}

\subsection{Models and datasets}

Our base models are DeepScaleR-1.5B-Preview~\cite{deepscaler2025} and DeepCoder-14B-Preview~\citep{deepcoder2025}, which are fine-tuned from DeepSeekR1-Distill-Qwen-1.5B and 14B~\citep{deepseekai2025deepseekr1incentivizingreasoningcapability} through iterative context lengthening. For training data, we follow the same datasets used in~\cite{deepscaler2025,deepcoder2025}. In the math domain, the training set consists of AIME (1984-2023), AMC, Omni-Math~\citep{gao2024omnimathuniversalolympiadlevel}, and STILL~\citep{min2024imitateexploreselfimprovereproduction}. For code training, we use TACO~\citep{li2023taco}, SYNTHETIC-1~\citep{2025synthetic1}, and LiveCodeBench (2023/05/01-2024/07/31)~\citep{jain2024livecodebench}. For evaluation, we use AIME 2024, MATH500~\citep{hendrycks2021measuringmathematicalproblemsolving}, AMC, Olympiad-Bench~\citep{gao2024omnimathuniversalolympiadlevel}, and Minerva Math~\cite{lewkowycz2022solving} for mathematical reasoning. For code-related tasks, we evaluate on LiveCodeBench (2024/08/01-2025/02/01)~\citep{jain2024livecodebench}, Codeforces, and HumanEval+~\cite{evalplus}. More training details are in Appendix~\ref{sec:detail}.

\subsection{Mathematical reasoning rsults}
We visualize the reward and validation Pass@1 performance on AIME2024 every 10 steps during training in Figure~\ref{fig:reward}. It can be observed that the reward steadily increases during the initial training phase and begins to converge after approximately the 150\textsuperscript{th} step. Meanwhile, the validation accuracy (Pass@1) improves rapidly, rising from around 0.07 to 0.20 over the course of training. This demonstrates that, through budget-constrained rollout, the model can quickly learn to reason effectively when the thinking phase is incomplete.

We report Pass@1 accuracy versus the number of tokens used across five math benchmarks AIME, AMC, Olympiad-Bench, MATH500, and Minerva Math in Figure~\ref{fig:math}. Our proposed method, E1-Math-1.5B, under both budget-constrained and 24K-token settings (red stars), consistently outperforms S1 (Budget Forcing) and L1-Exact, and performs competitively with L1-Max, while requiring significantly fewer training steps. On MATH500, E1-Math-1.5B achieves a Pass@1 accuracy of 83.6\% using only 1619 tokens per question, whereas L1-Exact and L1-Max yield lower or comparable performance with more tokens (L1-Exact: 79.9\% with 1959 tokens; L1-Max: 83.6\% with 1796 tokens). Notably, when evaluated without inference-time budget constraints, E1-Math-1.5B achieves higher accuracy than all baseline methods across all benchmarks. For example, on AIME2024, E1-Math-1.5B exhibits a performance degradation of only 6.0\% relative to the original model, compared to 12.9\% for L1-Max and 16.8\% for L1-Exact. These results demonstrate that our method is not only effective in enforcing inference-time budget constraints but also preserves most of the original model's performance. When compared with the original DeepScaleR-1.5B, E1-Math-1.5B reduces the average number of tokens used across datasets by more than 30\%, including a 32.1\% reduction on AIME2024.

Furthermore, similar to L1, S1, and O1, we observe a clear log-linear scaling pattern in E1: performance improves approximately linearly with respect to the logarithm of the number of generated reasoning tokens.

\subsection{Code reasoning results}

\begin{wrapfigure}{r}{0.5\textwidth}
\vspace{-1em}
    \centering
    \includegraphics[width=0.5\textwidth]{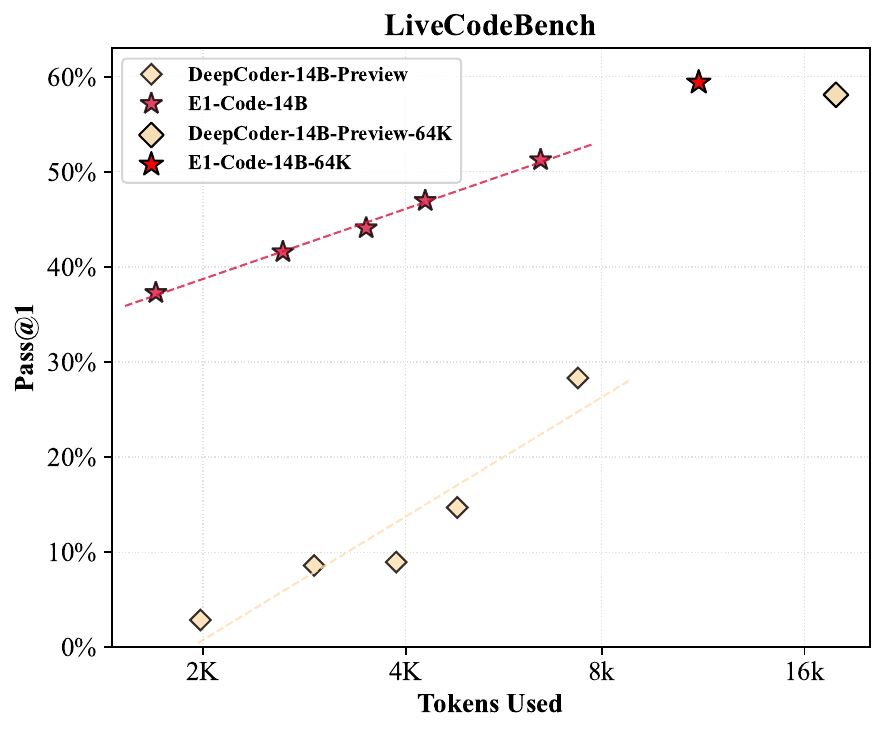}

    \vspace{-1em}
    \caption{Pass@1 accuracy on LiveCodeBench under varying reasoning budgets. Both of the models inference with separate budgeting.}
    \label{fig:livecode}
\vspace{-1em}
\end{wrapfigure}

As shown in Figure~\ref{fig:livecode}, we visualize the Pass@1 accuracy on LiveCodeBench under varying generation budgets, comparing our method to a simple separate budgeting strategy for thinking and solution. We observe that the original \textbf{DeepCoder-14B-Preview} fails to generate correct outputs when reasoning is incomplete, consistently achieving less than 10\% accuracy when inference budget is less than 4K even using separate budgeting. In contrast, our E1-Code-14B model demonstrates impressive scalability: its performance improves steadily as the inference budget increases, highlighting the effectiveness of our training strategy in enabling the model to reason adaptively under constrained thinking. Notably, E1-Code-14B also achieves a performance improvement of 0.3\% on LiveCodeBench even in the unconstrained setting, while simultaneously reducing the average number of generated tokens by \textbf{37.4\%}—from 17{,}815 to 11{,}145 tokens. This indicates that our method not only scales well with inference budgets but also promotes more concise and efficient reasoning.

\begin{table}[t]
\centering
\scriptsize
\caption{Comparison of models across LiveCodeBench, Codeforces, HumanEval+, and AIME benchmarks. E1-code-14B variants trained exclusively on code data; their AIME scores, obtained on math problems unseen during training, demonstrate that E1-code-14B retains strong math performance.}
\vspace{0.1cm}
\begin{tabular}{l|cccc|c}
\toprule
\textbf{Model} & \textbf{LiveCodeBench} & \textbf{Codeforces Rating} & \textbf{Codeforces Percentile} & \textbf{HumanEval+} & \textbf{AIME} \\
\midrule
O1-2024-12-17 (Low)              & 59.5          & \textbf{1991} & \textbf{96.1} & 90.8 & \textbf{74.4} \\
O3-Mini-2025-1-31 (Low)          & \textbf{60.9} & 1918 & 94.9 & \textbf{92.6} & 60.0 \\
O1-Preview                       & 42.7          & 1658 & 88.5 & 89.0          & 40.0 \\
DeepSeek-R1                      & \textbf{62.8} & 1948 & 95.4 & \textbf{92.6} & \textbf{79.8} \\
\midrule
DeepSeek-R1-Distill-Qwen-14B     & 53.0          & 1791 & 92.7 & 92.0          & 69.7 \\
DeepCoder-14B-Preview\tablefootnote{Results are reproduced using the authors' official code and model with the same evaluation protocol.}    & 58.1 & 1945 & 95.4 & 90.8 & 71.7 \\
\rowcolor{gray!20}  E1-code-14B $(t=1k, a=1k)$ & 37.3 & 1457 & 78.1 & 88.3 & 17.9\\
\rowcolor{gray!20}E1-code-14B $(t=2k, a=1k)$ & 41.6 & 1604 & 85.4 & 89.6 & 28.5\\
\rowcolor{gray!20}E1-code-14B $(t=3k, a=1k)$ & 44.1 & 1711 & 90.6 & 90.8 & 35.4\\
\rowcolor{gray!20}E1-code-14B $(t=4k, a=1k)$ & 47.0 & 1771 & 92.3 & 92.0 & 41.9\\
\rowcolor{gray!20}E1-code-14B & 58.4$\mathbf{_{+0.3}}$ & \textbf{1987}$\mathbf{_{+42}}$ & \textbf{96.0}$\mathbf{_{+0.6}}$ & 91.4$\mathbf{_{+0.6}}$ & 70.6\\
\bottomrule
\end{tabular}
\label{tab:code}
\end{table}

In Table~\ref{tab:code}, we report the performance of E1-Code-14B on four benchmarks: LiveCodeBench, Codeforces, HumanEval Plus, and AIME2024. We observe consistent test-time scaling behavior across all benchmarks under constrained inference budgets. Beyond scalability, our model also demonstrates strong performance in the unconstrained setting. Specifically, we observe performance improvements on LiveCodeBench, Codeforces, and HumanEval Plus, and only a slight performance drop on AIME2024. On Codeforces, E1-Code-14B achieves a 42-point improvement in rating and a 0.6 percentile gain, outperforming O3-Mini-2025-1-31 (Low) and performing comparably to O1-2024-12-17 (Low). These results highlight that our method not only enables efficient, budget-constrained reasoning but also enhances overall reasoning capability, even in unconstrained scenarios.

\subsection{Analysis and discussions}
\subsubsection{Which part is enhanced after training?}

To better understand which components of the reasoning process are enhanced through training, we conduct ablation experiments on DeepScaleR-1.5B-Preview and E1-Math-1.5B using the AIME2024 benchmark. Specifically, we separately generate the \textit{thinking} and \textit{solution} segments using both models under varying generation budgets. For example, we use DeepScaleR-1.5B-Preview to generate the thinking part, and then use E1-Math-1.5B to generate the corresponding solution based on that reasoning. This setup allows us to isolate the contributions of each model to the reasoning pipeline and assess how training improves each component.

As shown in Table~\ref{tab:ablation-parts}, we observe that both the thinking and solution are enhanced after training. Notably, the improvement in the solution component is more substantial, particularly under constrained thinking budgets. For instance, using the E1 model to generate only the solution segment yields an 8.7\% gain in accuracy compared to using the original DeepScaleR model, under a generation budget of $(0.5\text{K} + 1\text{K})$ tokens. This highlights the effectiveness of budget-constrained rollout in strengthening the model’s ability to produce high-quality solutions based on incomplete reasoning.

This observation also helps explain why training with a fixed budget constraint (e.g., $(1\text{K}, 1\text{K})$) enables the model to generalize effectively to a wide range of budget configurations. We hypothesize that the improvement in solution generation plays a central role in this generalization, allowing the model to adapt even when the available thinking tokens are reduced.
\begin{table}[t]
\centering
\small
\vspace{-1em}
\caption{Ablation of enhanced thinking and solution on DeepScaleR-1.5B-Preview and E1-Math-1.5. Budget is in format `thinking+solution` (in thousands of tokens).}
\vspace{0.2cm}
\begin{tabular}{cccc|llll}
\toprule
\multicolumn{2}{c}{\textbf{DeepScaleR-1.5B}} & \multicolumn{2}{c|}{\textbf{E1-Math-1.5B}} & \multicolumn{4}{c}{\textbf{Pass@1 (\%)}} \\
\textbf{Thinking} & \textbf{Solution} & \textbf{Thinking} & \textbf{Solution} & \textbf{0.5\text{K}+1\text{K}} & \textbf{1\text{K}+1\text{K}} & \textbf{2\text{K}+1\text{K}} & \textbf{3\text{K}+1\text{K}} \\
\midrule
\cmark & \cmark & \xmark & \xmark & 2.10          & 4.80            & 12.5          & 20.0 \\
\xmark & \cmark & \cmark & \xmark & 3.50$_{+1.4}$ & 7.90$_{+3.1}$   & 20.6$_{+8.1}$ & 24.0$_{+4.0}$ \\
\cmark & \xmark & \xmark & \cmark & 10.8$_{+8.7}$ & 14.2$_{+9.4}$   & 21.9$_{+9.4}$ & 26.4$_{+6.4}$ \\
\xmark & \xmark & \cmark & \cmark & 13.5$\mathbf{_{+11.4}}$ & 17.5$\mathbf{_{+12.7}}$ & 24.8$\mathbf{_{+12.3}}$ & 27.9$\mathbf{_{+7.9}}$ \\
\bottomrule
\end{tabular}
\label{tab:ablation-parts}
\vspace{-1em}
\end{table}

\subsubsection{Ablation of training budget \texorpdfstring{$t^*$}{t*}}

\begin{figure}
    \centering
    \includegraphics[width=0.95\linewidth]{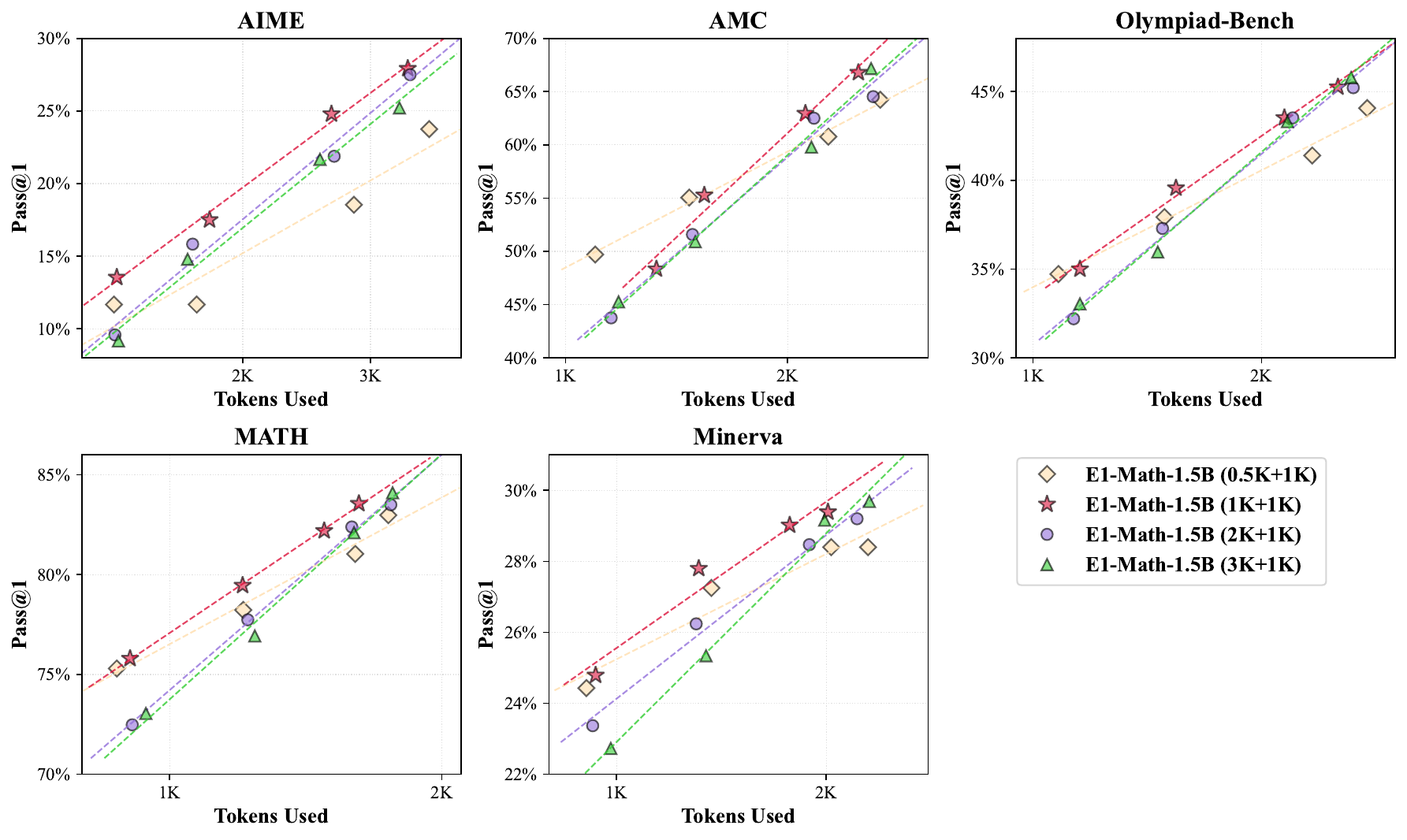}
    \caption{Ablation study on training reasoning budget \( t^* \). We compare four settings: \( t^* \in \{0.5\text{K},\ 1\text{K},\ 2\text{K},\ 3\text{K} \} \), while keeping the solution budget fixed at \( a^* = 1\text{K} \).}
    \label{fig:ablation}
    \vspace{-1em}
\end{figure}

To further investigate the role of the thinking budget $t^*$ in our proposed budget-constrained rollout, we conduct experiments to evaluate the model’s performance under four settings: $t^* \in \{0.5\text{K},\ 1\text{K},\ 2\text{K},\ 3\text{K}\}$, while keeping the solution budget fixed at $a^* = 1\text{K}$. We evaluate on five math benchmarks: AIME, AMC, Olympiad-Bench, MATH500, and Minerva Math (Figure~\ref{fig:ablation}).

Across all configurations, the model demonstrates strong generalization to varying inference budgets on all benchmarks. Among the tested values, $t^* = 1\text{K}$ consistently achieves the best performance, while also maintaining a low maximum generation length of 2K tokens, making it a highly efficient and effective setting. Based on this trade-off between performance and computational cost, we adopt $(t^*=1\text{K},\ s^*=1\text{K})$ as our default configuration.

\subsubsection{Token allocation between thinking and solution}

\begin{figure}[t]
    \centering
    \includegraphics[width=0.95\linewidth]{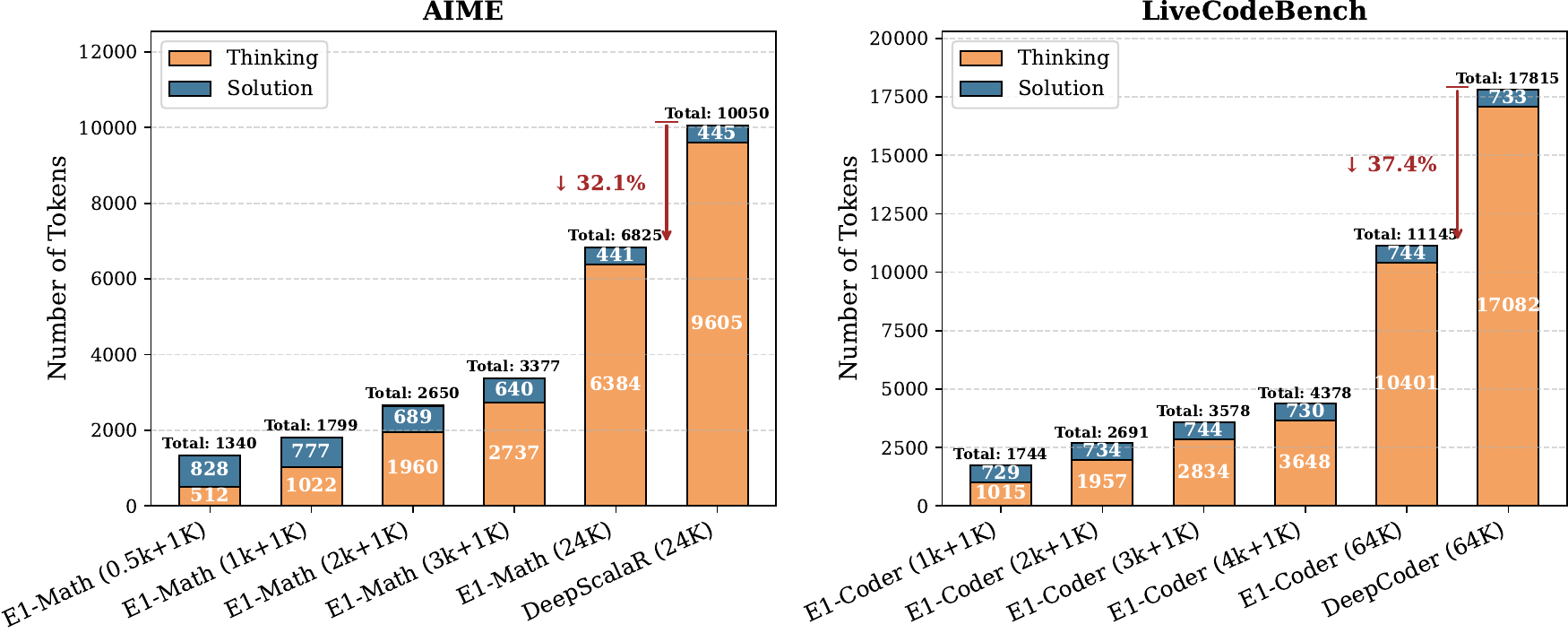}
    \caption{Distribution of tokens for thinking and solution across different generation budgets.}
    \label{fig:hist}
\end{figure}

Figure~\ref{fig:hist} visualizes the distribution of \textit{thinking} and \textit{solution} tokens within generated trajectories under different generation budget constraints. We select AIME2024 for the math task and LiveCodeBench for the coding task.

For AIME2024, as the inference budget decreases, the number of tokens used in the thinking segment decreases accordingly, while the number of tokens in the solution segment slightly increases. A similar trend is observed on LiveCodeBench, where the thinking tokens decrease with tighter budgets, while the number of solution tokens remains relatively stable.

Notably, even when evaluated without budget constraints, our trained E1 models demonstrate substantial token efficiency: they reduce total token usage by 32.1\% on AIME2024 and 37.4\% on LiveCodeBench, while maintaining strong performance (even slightly better than the baseline model). This suggests that the model has learned to reason more concisely and generate efficient solutions post training.

\subsubsection{Iterative training}
\begin{table}[h]
    \centering
    \caption{Pass@1(\%) on AIME 2024 across different budget configurations in two iterations.}
    \vspace{2mm}
    \small
    \begin{tabular}{c|c c c c}
    \toprule
    \textbf{Iteration} & \textbf{0.5K+1K} & \textbf{1K+1K} & \textbf{2K+1K} & \textbf{3K+1K} \\
    \midrule
    1\textsuperscript{st} & 13.5 & 17.5 & 24.8 & 27.9 \\
    2\textsuperscript{nd} & 11.5 & 17.1 & 22.9 & 26.7 \\
    \bottomrule
    \end{tabular}
    \label{tab:iterative}
\end{table}
In this section, we investigate whether the model can benefit from \textit{iterative training}—that is, performing a second round of training with expanded budgets after initial training. Specifically, we first train the model using a budget of $(t^* = 1\text{K},\ a^* = 1\text{K})$, and then continue training from the resulting checkpoint using a larger thinking budget of $(t^* = 3\text{K},\ a^* = 1\text{K})$. The results on AIME2024 are reported in Table~\ref{tab:iterative}.

Surprisingly, the second round of training does not improve performance. In fact, we observe a slight degradation in accuracy, suggesting that once the model has learned to reason effectively under a shorter budget, further training with a longer budget may not provide additional benefits.

\section{Conclusion}
We introduce \textbf{Elastic Reasoning}, a unified framework for enabling large reasoning models to generate accurate and efficient chain-of-thought outputs under strict inference-time constraints. By explicitly separating the reasoning process into \emph{thinking} and \emph{solution} phases, and training with a novel \emph{budget-constrained rollout} strategy, our approach ensures robustness to truncated reasoning while preserving or even improving overall performance. Elastic Reasoning significantly reduces token usage during inference, generalizes across unseen budget configurations, and outperforms prior length control baselines in both mathematical and programming domains. Our findings offer a scalable and principled solution for real-world deployment of reasoning LLMs where computation budgets are limited. We believe this framework opens new directions for budget-aware reasoning.

\bibliography{my}
\bibliographystyle{plainnat}

\appendix

\section{Limitations and broader impacts}
While Elastic Reasoning offers a practical and scalable solution for length-controllable inference, it introduces a few limitations. First, our method assumes that the reasoning process can be meaningfully segmented into distinct thinking and solution phases, which may not hold for tasks with highly interleaved reasoning and answering (e.g., some forms of dialogue or commonsense reasoning). Second, our evaluation focuses primarily on math and code reasoning tasks; further investigation is needed to assess the effectiveness of Elastic Reasoning in domains such as scientific question answering or multi-hop retrieval.

Elastic Reasoning aims to improve the deployability and controllability of large reasoning models by enabling efficient inference under strict compute constraints. This can have positive downstream effects, such as reducing energy consumption, lowering latency for interactive applications, and enabling access to reasoning-capable models in resource-limited environments. At the same time, the ability to tailor model outputs based on budget constraints introduces new challenges: models may be optimized to generate shorter outputs without transparent communication of omitted reasoning, potentially affecting interpretability or user trust. Additionally, as reasoning models are increasingly used in high-stakes domains (e.g., education, law, healthcare), the separation between “thinking” and “solution” phases may require careful design to ensure that important justifications are not lost during inference-time truncation. We encourage future work to explore human-centered evaluations of controllable reasoning, and to develop standards for safely deploying budget-aware models in real-world applications.

\section{Training details}~\label{sec:detail}
For GRPO training, we adopt the same hyperparameters as those used in DeepScaleR-1.5B-Preview and DeepCode-14B-Preview. For E1-Math-1.5B, we use a learning rate of \(1 \times 10^{-6}\) and a batch size of 128. The maximum context length is set to 1K tokens for the thinking segment and 1K tokens for the solution segment. Training is performed for 200 steps using the \texttt{VeRL}~\citep{sheng2024hybridflow} framework. For E1-Code-14B, we use the same learning rate and batch size. The context length configuration mirrors that of the math model: 1K tokens for thinking and 1K tokens for solution. Training is conducted for only 30 steps.


\end{document}